\documentclass[letterpaper, 10 pt, conference]{ieeeconf}  

\IEEEoverridecommandlockouts                              

\usepackage{graphicx}      
\usepackage{amsmath} 
\usepackage{amssymb}  
\usepackage{multirow}
\usepackage{float}
\usepackage{algorithm}
\usepackage{subcaption}
\usepackage{caption}
\usepackage{wrapfig}
\newtheorem{defin}{Definition}
\newtheorem{lemma}{Lemma}
\newtheorem{thm}{Theorem}
\newtheorem{cor}{Corollary}
\title{\LARGE \bf
Hilbert's Space-filling Curve for Regions with Holes
}

\author{Siddharth H. Nair$^{1}$, Arpita Sinha$^{2}$ and Leena Vachhani$^{3}$
\thanks{$^{1}$Siddharth H. Nair is with Department of Aerospace Engineering, 
   Indian Institute of Technology Bombay, 400076 India.        {\tt\small siddharth.nair@iitb.ac.in}}%
\thanks{$^{2}$Arpita Sinha is with the Faculty of Systems and Control Engineering, 
   Indian Institute of Technology Bombay, 400076 India.  {\tt\small arpita.sinha@iitb.ac.in}}%
\thanks{$^{3}$Leena Vachhani is with the Faculty of Systems and Control Engineering, 
   Indian Institute of Technology Bombay, 400076 India.  {\tt\small leena.vachhani@iitb.ac.in}}%
}

\begin{document}

\maketitle
\thispagestyle{empty}
\pagestyle{empty}

\begin{abstract}
The paper presents a systematic strategy for implementing Hilbert's space filling curve for use in online exploration tasks and addresses its application in scenarios wherein the space to be searched obstacles (or holes) whose locations are not known a priori. Using the self-similarity and locality preserving properties of Hilbert's space filling curve, a set of evasive maneuvers are prescribed and characterized for online implementation. Application of these maneuvers in the case of non-uniform coverage of spaces and for obstacles of varying sizes is also presented. The results are validated with representative simulations demonstrating the deployment of the approach.
\end{abstract}

\section{INTRODUCTION}
Investigations on space filling curves began when George Cantor proved that there exists a bijective map between any two finite-dimensional manifolds. The continuity of such maps, however, was a mystery until E. Netto demonstrated that a bijective map between two finite-dimensional manifolds of different dimensions is necessarily discontinuous. All was not lost though- in 1890, Peano relaxed the bijectivity condition of such maps and discovered a continuous surjective mapping from the unit interval $\mathbf{I}=[0,1]$ to the unit square $\mathbf{Q}=[0,1]^{2}$. This paved the way to designing continuous maps to fill a higher dimensional space from a lower dimensional space\cite{c10}.\\

Space filling curves are primarily used in applications that require visiting various regions in a space. Numerous works pertaining to use of various space filling curves for optimization exist in literature (\cite{c1,c2,c3}). The advantage of using Hilbert curve lies within the very nature of the map. The images of points in the domain of the map neighbour each other thereby lending a sense of ``anticipation" during implementation. This inspires the use of the curve for robotic applications. In \cite{c4}, a robot motion planning problem is introduced and uses the Hilbert space-filling curve to uniformly cover a space with a single agent or multiple agents coordinating with each other. In \cite{c5}, space filling curves are used to decompose an arbitrary shape using space filling curves and thus use the technique for planning a path for a tool to machine the desired shape. In \cite{c6}, the strategy presented in \cite{c4} is implemented using an aerial robot for search operations and also uses Moore's space filling curve (a modification of the Hilbert curve). In \cite{c7}, a tree is used to arrange nodes for uniformly covering the unit square. The nodes are numbered according to that of the Hilbert curve. In \cite{c8}, a new approach is proposed to search this coverage tree to explore a space non-uniformly via an aerial agent by using the self-similarity of the Hilbert curve (which is a fractal curve) to define various levels for the tree.\\

Literature pertaining to exploration for robots using space filling curves in spaces with holes/obstacles is sparse. A rigorous solution to search a space with holes is offered in \cite{c9} with applications in sensor networks by mapping any search domain canonically to a torus and design space-filling trajectories on this torus. However, the method requires knowledge of the structure of the search domain beforehand and involves complex computations. An investigation of a technique that uses feedback and traces Hilbert's curve would support sensor based autonomous strategies. Therefore, this paper addresses tracing the Hilbert curve when holes of fixed size in the space are present, but their locations are sensed online. A modification of the map of the Hilbert's curve is also presented for robotic exploration applications. It shall be seen that a simplified approach can be derived by examining the properties of the Hilbert curve and its map. Furthermore, this is extended to the case of non-uniform coverage of spaces along the lines of \cite{c8} and for obstacles of varying sizes.\\

The rest of the paper is divided into five sections. Section \ref{S2} introduces the map that defines the Hilbert curve. Section \ref{S3} highlights the underlying properties of the Hilbert curve and addresses the obstacle avoidance problem while Section \ref{S4} extends the formulation for non-uniform coverage of spaces. Section \ref{S5} presents an algorithm for online implementation of the prescribed strategy with simulation results followed by concluding remarks in Section \ref{S6}.

\section{Preliminary on Hilbert's Space-filling Curve}\label{S2}
\begin{figure}[H]
\centering
\includegraphics[scale=0.4]{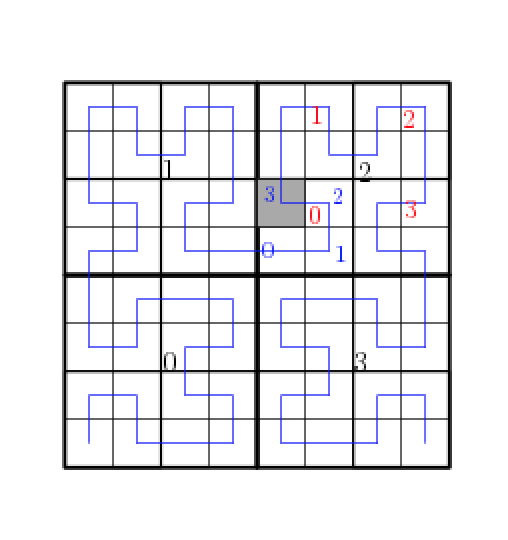}
\caption{$t=0._{4}203$ mapped into the unit square via Hilbert's map}
\label{ex1}
\end{figure}

\begin{figure}[H]
\centering
\begin{subfigure}{.25\textwidth}
  \centering
  \includegraphics[scale=0.2]{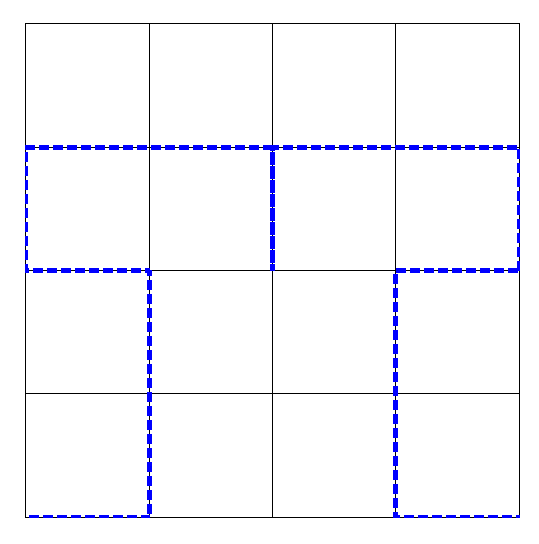}
  \caption{}
  \label{fig:old}
\end{subfigure}\begin{subfigure}{.25\textwidth}
  \centering
  \includegraphics[scale=0.3]{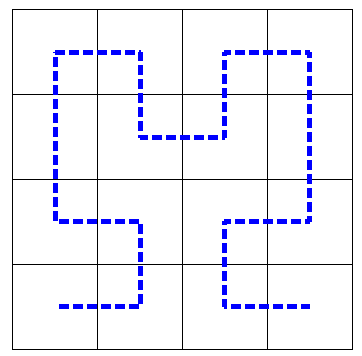}
  \caption{}
  \label{fig:new}
\end{subfigure}
\caption{Figures \ref{fig:old} and \ref{fig:new} are Hilbert's curves obtained using  (\ref{mapold}) and (\ref{mapnew}) respectively.}
\end{figure}

Hilbert \cite{c10} was the first to propose a geometric generation principle for the construction
of a space filling curve which can be encapsulated by the following procedure:
\begin{itemize}
\item The unit interval $\mathbf{I}$ is mapped continuously onto the unit-square $\mathbf{Q}$. $\mathbf{I}$ is partitioned into four equal sub-intervals and $\mathbf{Q}$ is partitioned into four congruent sub-squares, so as to map the same continuously onto one of the sub-squares. This is repeated for the all the sub-intervals and sub-squares.\\
\item When repeating this procedure ad infinitum, the sub-squares are arranged in such a way that adjacent sub-squares correspond to
adjacent sub-intervals hence preserving the overall continuity of the mapping.
\end{itemize}
Hence, the $nth$ order Hilbert curve divides the space $\mathbf{Q}$ and interval $\mathbf{I}$ into $4^{n}$ subsets.
Now, the map $f_{h}: \mathbf{I} \rightarrow \mathbf{Q}$ is defined such that every $t\in \mathbf{I}$ corresponds to a unique sequence of nested closed squares that shrink into a point of $\mathbf{Q}$, the image $f_{h}(t)$. Let us represent $t \in \mathbf{I}$ in quaternary form as $ t= 0._{4}q_{1}q_{2}q_{3}q_{4}....= \frac{q_{1}}{4} + \frac{q_{2}}{4^{2}} + \frac{q_{3}}{4^{3}} +...$ where $q_i=\{0,1,2,3\}$. The map $f_{h}: \mathbf{I} \rightarrow \mathbf{Q}$ is called the Hilbert space-filling curve and is defined as a composition of transformations as $ f_{h}(t) = \lim_{n\rightarrow \infty} T_{q_{1}}T_{q_{2}}..T_{q_{n}}\mathbf{Q}$ where the transformations $T_{q_i}$ acting on $\mathbf{Q}$ are defined as
\small
\begin{align*} 
T_{0}\left(\begin{bmatrix} x\\y \end{bmatrix}\right)&= \dfrac{1}{2}\underbrace{\begin{bmatrix}
0 & 1\\1 & 0 \end{bmatrix}}_{H_0} \begin{bmatrix} x\\y \end{bmatrix} + \dfrac{1}{2}\underbrace{\begin{bmatrix}0\\0\end{bmatrix}}_{h_0}\\
T_{1}\left(\begin{bmatrix} x\\y \end{bmatrix}\right)&= \dfrac{1}{2}\underbrace{\begin{bmatrix}
1 & 0\\0 & 1 \end{bmatrix}}_{H_1} \begin{bmatrix} x\\y \end{bmatrix} + \dfrac{1}{2}\underbrace{\begin{bmatrix}0\\1\end{bmatrix}}_{h_1}\\
T_{2}\left(\begin{bmatrix} x\\y \end{bmatrix}\right)&= \dfrac{1}{2}\underbrace{\begin{bmatrix}
1 & 0\\0 & 1 \end{bmatrix}}_{H_2} \begin{bmatrix} x\\y \end{bmatrix} + \dfrac{1}{2}\underbrace{\begin{bmatrix}1\\1\end{bmatrix}}_{h_2}\\
T_{3}\left(\begin{bmatrix} x\\y \end{bmatrix}\right)&= \dfrac{1}{2}\underbrace{\begin{bmatrix}
0 & -1\\-1 & 0 \end{bmatrix}}_{H_3} \begin{bmatrix} x\\y \end{bmatrix} + \dfrac{1}{2}\underbrace{\begin{bmatrix}2\\1\end{bmatrix}}_{h_3}\\
\end{align*}
\normalsize
For example, the image of the point $t=0._{4}203$ lies in the $(3+1)th$ sub-square, of the $(0+1)th$ sub-square, of the $(2+1)th$ sub-square of $\mathbf{Q}$ (see Figure \ref{ex1}). 
Also note that any finite quaternary represents the starting of an interval and so,
\begin{align}\label{mapold}
f(0._{4}q_{1}q_{2}..q_{n})&=T_{q_{1}}..T_{q_{n}}(\lim_{p\rightarrow\infty}T^p_{0}(\mathbf{Q}))\nonumber\\&=T_{q_{1}}..T_{q_{n}}\begin{bmatrix}0\\0\end{bmatrix}
\end{align}
Defining $e_{0j}= {No.\ of\ times\ T_{0}\ occurs\ before\ T_{q_{j}}\  (mod 2)}$ and $e_{3j}= {No.\ of\ times\ T_{3}\ occurs\ before\ T_{q_{j}}\  (mod 2)}$, Equation \eqref{mapold} simplifies (see \cite{c10}) to $$f(0._{4}q_{1}q_{2}..q_{n})= \sum_{j=1}^{n}\frac{1}{2^j}H_{0}^{e_{0j}}H_{3}^{e_{3j}}h_{q_{j}}$$ The implementation of a Hilbert curve is done via its approximating polygon which is obtained by mapping $4^n$ nodes of the form $ 0,\frac{1}{4^n},\frac{2}{4^n},...\frac{4^n-1}{4^n}$ to $\mathbf{Q}$ and joining the images using straight lines. Details of this analysis can be found in \cite{c10}. An implementation of a second order Hilbert curve is shown in Figure \ref{fig:old}.

For our application, we shift the nodes to the centres of the sub-squares and thus modify the map slightly as follows.
\small
\begin{align}\label{mapnew}
f(0._{4}q_{1}q_{2}..q_{n})&=T_{q_{1}}T_{q_{2}}..F_{q_{n}}=\frac{1}{2}H_{q_{1}}..H_{q_{n-1}}F_{q_{n}}+\sum_{j=1}^{n-1}\frac{1}{2^j}H_{0}^{e_{0j}}H_{3}^{e_{3j}}h_{q_{j}}
\end{align}
\normalsize
where $F_{0}=\begin{bmatrix}
\frac{1}{4}\\\frac{1}{4}
\end{bmatrix}$, $F_{1}=\begin{bmatrix}
\frac{1}{4}\\\frac{3}{4}
\end{bmatrix}$, $F_{2}=\begin{bmatrix}
\frac{3}{4}\\\frac{3}{4}
\end{bmatrix}$ and $F_{3}=\begin{bmatrix}
\frac{3}{4}\\\frac{1}{4}
\end{bmatrix}$.\\\\
An implementation of the modified map is shown in Figure \ref{fig:new}. This modification is motivated for exploration tasks where it would make sense to identify each node (sub-square) in the search space by its centre and not by its edges which could be shared with other nodes.\\

This implementation of the Hilbert's curve is restricted to all euclidean spaces that are homeomorphic to a unit square (polygons, circles etc.). The nodes obtained for the unit square can be mapped back into the actual space via the homeomorphism (\cite{c13}).

\addtolength{\textheight}{-3cm}   

\section{Extension of Hilbert's Space-filling curve to domains with blocked nodes}\label{S3}
Consider a problem wherein a searching agent traverses the space $\mathbf{Q}$ along the nodes of the $nth$ order Hilbert curve and there is a single "obstacle" node of the same resolution. The presence of the obstacle leaves that particular node unaccessible or blocked. The space is "unknown" in the sense that the searching agent starts from a corner of the search space and only the boundary of the search space is known. It is also assumed that information about the occupancy of the neighbouring nodes (the sub-squares touching the sub-square occupied by the agent) is available to the agent. Before introducing the obstacle avoidance strategy, some definitions and properties of the Hilbert curve are presented.

\begin{defin}
   The nodes that enter or leave one of the $(n-1)th$ order Hilbert curves in the $nth$ order Hilbert curve are defined as corner nodes of the $nth$ order Hilbert curve.
   \label{def}
\end{defin}

\begin{figure}[h]
\centering
\hspace{0.5cm}
\includegraphics[scale=0.3]{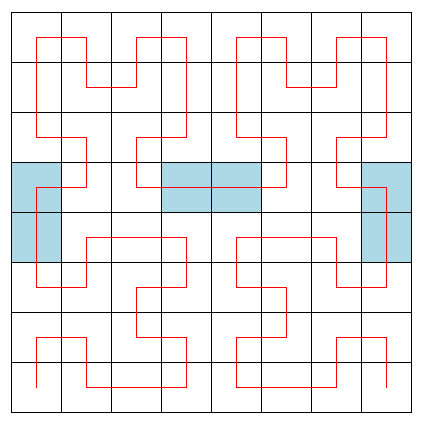}
\caption{Corner points for a third order Hilbert curve}
\end{figure}

\begin{figure}[h]
\centering
\begin{subfigure}{.25\textwidth}
\captionsetup{width=0.8\textwidth}
  \centering
  \includegraphics[scale=0.47]{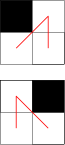}
  \caption{\small{Evasive maneuver for when the obstacle lies on a non-corner node}}
\label{nc}
\end{subfigure}\begin{subfigure}{.25\textwidth}
\captionsetup{width=0.8\textwidth}
  \centering
  \includegraphics[scale=0.35]{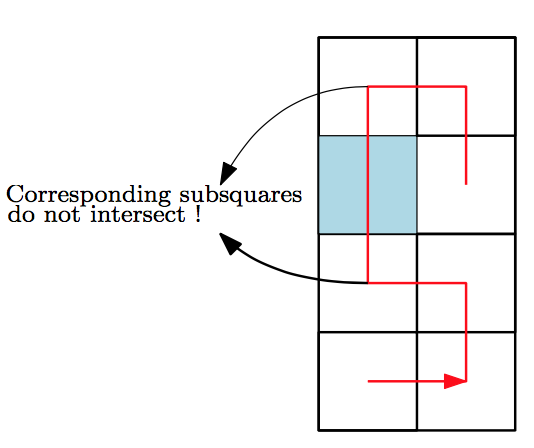}
  \caption{A case where skipping the blocked node is not possible.}
\label{exx}
\end{subfigure}
\caption{}
\end{figure}

\begin{lemma}
The corner points of a Hilbert curve of order $n$ are given by $t=0._{4}q_{1}\underbrace{00...0}_{n-1\ 0s}$ or $t=0._{4}q_{1}\underbrace{33...3}_{n-1\ 3s}$.\\
\end{lemma}
\textit{Proof }
By definition, intervals of the form $[\frac{q_{1}}{4},\frac{q_{1}+1}{4}]$ map into one of the four $(n-1)th$ order Hilbert curves in the $nth$ order Hilbert curve. Therefore, the corner node that enters the $(n-1)th$ order Hilbert curve is simply the first of the $4^{n-1}$ nodes within the same. Hence, it is given by $t=0._{4}q_{1}00...0$.\\
Similarly, the corner node that leaves the $(n-1)th$ order Hilbert curve is the last node of the same and is given by $t=0._{4}q_{1}33...3$.\hspace{6cm}$\blacksquare$\\\\
Note that each lower order Hilbert curve within the $nth$ order Hilbert curve also contains corner nodes defined in a similar fashion. In general, each corner node (of every order) is given by $t=0._{4}q_{1}q_{2}..q_{n}$ where $q_{n}=0$ or $3$. Hence, all nodes that enter or leave the first order Hilbert curve are the corner nodes. Every other node lies between two corner nodes and due to the locality preserving property of the Hilbert curve, lie adjacent to each other. This is an important consequence which helps in devising a simple strategy to avoid an obstacle that is placed on a non-corner node as shown in Figure \ref{nc} which depicts a first order Hilbert curve.
In essence, the locality preserving property (\cite{c12}) of the Hilbert curve helps us to skip the blocked node. However if the blocked node is a corner node, it may not be possible to simply skip it because the sub-square containing the next node may not intersect the sub-square containing the node preceding the obstacle node. Figure \ref{exx} illustrates the point in the case of a second order Hilbert curve where simply skipping the blocked node is not possible without disturbing the sequence of nodes dictated by map for the Hilbert curve.

Now, we characterize the factors that come into play when deciding on a maneuver to avoid an obstacle node when placed on a corner node for a Hilbert curve of any order. Additionally, we also assume that the first node and the last node of the $nth$ order Hilbert curve are unblocked.\\
\begin{thm}
For the $nth$ order Hilbert curve, let $e_{n}= n\ (mod\ 2)$ and the entering and exiting corner nodes be given by $t=0._{4}(q_{1}+1)00...0$ and $t=0._{4}q_{1}33...3$ respectively where $q_{1}=0,1,2$. Then, the required evasive maneuver to avoid an obstacle on a corner node depends solely on the type of corner node  (i.e,  entering node or exiting node).\\
\end{thm}
\textit{Proof }
Consider the corner node which leaves the $(q_{1}+1)$th sub-square of the $n$th order Hilbert curve. It is denoted by $t=0._{4}q_{1}33...3$. The succeeding and preceding nodes are $t_{1}=0._{4}(q_{1}+1)00...0$ and $t_{2}=0._{4}q_{1}33...2$. These points are mapped into $\mathbf{Q}$ as follows.
\small
\begin{align*}f_{h}(t_{1})=&\frac{1}{2^{n-1}}H_{q_{1}+1}H_{0}H_{0}...F_{0}+\frac{1}{2}h_{q_{1}+1}+\\ &\frac{1}{2^{2}}H_{q_{1}+1}(h_{0}+\frac{1}{2}H_{0}h_{0}+\frac{1}{2^{2}}h_{0}+..\frac{1}{2^{n-1}}H_{0}^{1-e_{n}}h_{0})\end{align*}
\normalsize
Since $h_{0}=\begin{bmatrix} 
0\\0 
\end{bmatrix}$ and $F_{0}=\begin{bmatrix}\frac{1}{4}\\\frac{1}{4}\end{bmatrix}$, we have
\small
$$f_{h}(t_{1})=\frac{1}{2^{n-1}}H_{q_{1}+1}F_{0}+\frac{1}{2}h_{q_{1}+1}$$
\normalsize
Now,
\small
\begin{align*}f_{h}(t_{2})=&\frac{1}{2^{n-1}}H_{q_{1}}H_{3}H_{3}...F_{2}+\frac{1}{2}h_{q_{1}}+\\ &\frac{1}{2^{2}}H_{q_{1}}(h_{3}+\frac{1}{2}H_{3}h_{3}+\frac{1}{2^{2}}h_{3}+..\frac{1}{2^{n-1}}H_{3}^{1-e_{n}}h_{3})\end{align*}
\normalsize
Again, since $H_{3}=-H_{0}$ and $F_{2}=\begin{bmatrix}\frac{3}{4}\\\frac{3}{4}\end{bmatrix}$, we have
\small
\begin{align*}f_{h}&(t_{2})=-\frac{1}{2^{n-1}}H_{q_{1}}F_{2}+\frac{1}{2}h_{q_{1}}+\\ &\frac{1}{2^{2}}H_{q_{1}}(h_{3}-\frac{1}{2}H_{0}h_{3}+\frac{1}{2^{2}}h_{3}+..\frac{1}{2^{n-1}}(-H_{0})^{1-e_{n}}h_{3})\end{align*}

\begin{align*}\Rightarrow f_{h}&(t_{2})=-\frac{1}{2^{n-1}}H_{q_{1}}F_{2}+\frac{1}{2}h_{q_{1}}+\\ &\frac{1}{2^{2}}H_{q_{1}}(h_{3}-\frac{1}{2}H_{0}h_{3}+\frac{1}{2^{2}}h_{3}+..\frac{1}{2^{n-1}}(-H_{0})^{1-e_{n}}h_{3}) \\\end{align*}
\normalsize

The evasive maneuver to be chosen depends on the distance between the nodes preceding and succeeding the obstacle node. Hence, we find $f_h(t_{1})-f_h(t_{2})$.
\small
\begin{align*}f_h(t_{1})-&f_h(t_{2})=\frac{1}{2^{n-1}}(H_{q_{1}+1}F_{0}+H_{q_{1}}F_{2})+\frac{1}{2}(h_{q_{1}+1}-h_{q_{1}})\\ &-\frac{1}{2^2}H_{q_{1}}\begin{bmatrix} 2-\frac{1}{2}+\frac{1}{2}-\frac{1}{4}+\frac{1}{4}..-\frac{1}{2^{n-3}}(1-e_{n})\\1-1+\frac{1}{4}-\frac{1}{4}+\frac{1}{8}-\frac{1}{8}-\frac{1}{2^{n-2}}e_{n} \end{bmatrix}\end{align*}
\normalsize
Defining $e_{q_{1}}= q_{1}\ (mod 2)$,
\small
\begin{align*}\Rightarrow f_h(t_{1})-f_h(t_{2})=&\begin{bmatrix} (\frac{1}{2^{n}})^{floor(\frac{q_{1}}{2})}+\frac{e_{q_{1}}}{2}\\\frac{1}{2^{n}})^{floor(\frac{q_{1}}{2})}+\frac{1-e_{q_{1}}}{2} \end{bmatrix}\\ &-\frac{1}{2^2}H_{q_{1}}\begin{bmatrix} 2-\frac{1}{2^{n-3}}(1-e_{n})\\ -\frac{1}{2^{n-2}}e_{n}\end{bmatrix}\end{align*}
\normalsize

Noting the fact that term $\frac{1}{2^{n}}$ is merely a scaling factor, the distant between the nodes, $||f_h(t_{1})-f_h(t_{2})||_{2}$, is primarily dependent on $q_{1}$ and $e_{n}$. A similar analysis for the second type of corner node(entering a Hilbert curve of the same order) yields a similar result.\hspace{4.1cm}$\blacksquare$\\\\
The same result can be obtained for lower order curves within the given Hilbert curve because of its self-similarity property.  Consider a general corner node $t=0._{4}q_{1}q_{2}..q_{p}mmm..m$ where $m$ is either 0 or 3. This node enters or exits an $(n-p)th$ order Hilbert curve and so, the "effective" order of the Hilbert curve is $n_{eff}=n-p+1$. So, to decide on an evasive maneuver, we obtain $e_{n_{eff}},\ q_{p},$ and $m$ (which is known to the agent due to position feedback). Table \ref{tabev} depicts the various evasive maneuvers that can be employed for various scenarios to avoid an obstacle placed on a corner node. By identifying the dependancy of the maneuver on $e_{n},\ q_{p},$ and $m$, the various scenarios can be identified canonically and hence, presents the possibility of designing strategies for evasion that can be used online. Considering the assumption that the first node and the last node of the $nth$ order Hilbert curve are unblocked, the number of combinations of $(e_{n},q_p,m)$ is given by $$\underbrace{2}_{e_n}\times \underbrace{4}_{q_p} \times \underbrace{2}_{m} - \underbrace{4}_{Start\ and\ end\ points}=12$$  
The various combinations are categorized together based on the distance between the current node and the node succeeding the obstacle.
The evasive strategies are presented in two forms :\begin{itemize}
\item in a discrete setting wherein the strategy prescribes the sequence of nodes to be traversed to evade the obstacle and not disturb the sequence of nodes as prescribed by Hilbert's map.
\item in a continuous setting wherein the piecewise linear map that defines the approximating polygon for the Hilbert curve is modified.
\end{itemize}

For example, for a third order Hilbert curve if the position of the obstacle is $t=0._{4}110$, then $p=2$ and $n_{eff}=3-2+1=2$. Therefore, $\ e_{n_{eff}}=0, \ q_{p}=1,\ m=0$ and the prescribed maneuver is the last maneuver of the table. The paths described in Table \ref{tabev} can be implemented using mobile robots by fitting splines along the nodes as is shown in \cite{c11}

\begin{table*}
\centering
  \begin{tabular}
      {llll} \hline\hline $(e_{n},q_p,m)$ & Evasion Maneuver & Change in Path &  Change in Map\\
      & & & $f':\mathbf{I} \rightarrow \mathbf{Q}$\\
    \hline\hline \begin{tabular}{@{}c@{}}(0,0,3)\\(0,3,0)\\(1,1,0)\\(1,2,3)
      \end{tabular}& \parbox[c]{2em}{
      \includegraphics[scale=0.4]{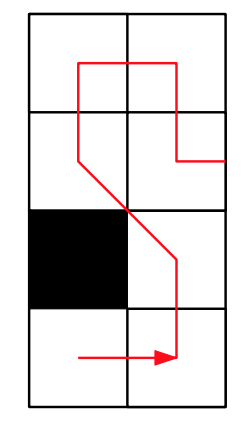}} & \multirow{2}{*}{\begin{tabular}{@{}c@{}}\\ \\ \\ \\ Node$(:,n_{obs})=$ Node$(:,n_{obs}+1)$\end{tabular}} & \multirow{2}{*}{ \begin{tabular}{@{}c@{}} \\ \\ \\ \\$(t-\frac{n_{obs}-1}{4^{n}})f_{h}(\frac{n_{obs}+1}{4^{n}})-(t-\frac{n_{obs}+1}{4^{n}})f_{h}(\frac{n_{obs}-1}{4^{n}})$\\ $\forall t \in [\frac{n_{obs}-1}{4^{n}},\frac{n_{obs}+1}{4^{n}}]$\end{tabular}
}\\
      \begin{tabular}{@{}c@{}}(0,1,3)\\(0,2,0) 
      \end{tabular}& \parbox[c]{2em}{
      \includegraphics[scale=0.4]{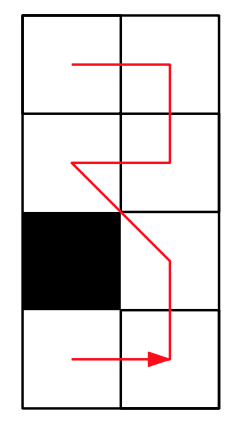}}&&\\ \hline
      
\begin{tabular}{@{}c@{}}(1,0,3)\\(0,2,3) 
\end{tabular} & \parbox[c]{2em}{
\includegraphics[scale=0.4]{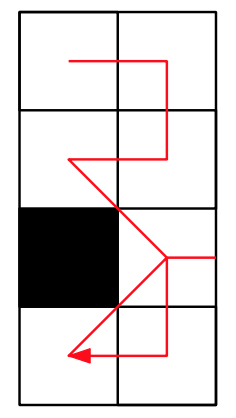}} &\multirow{2}{*}{\begin{tabular}{@{}c@{}}\\ \\ \\ Node$(:,n_{obs})=$ Node$(:,n_{obs}-3)$\end{tabular}}& \multirow{2}{*}{\begin{tabular}{@{}c@{}} \\$(t-\frac{n_{obs}-1}{4^{n}})f_{h}(\frac{n_{obs}-3}{4^{n}})-(t-\frac{n_{obs}}{4^{n}})f_{h}(\frac{n_{obs}-1}{4^{n}})$\\ $\forall t \in [\frac{n_{obs}-1}{4^{n}},\frac{n_{obs}}{4^{n}}]$\\\\$(t-\frac{n_{obs}}{4^{n}})f_{h}(\frac{n_{obs}+1}{4^{n}})-(t-\frac{n_{obs}+1}{4^{n}})f_{h}(\frac{n_{obs}-3}{4^{n}})$\\ $\forall t \in [\frac{n_{obs}}{4^{n}},\frac{n_{obs}+1}{4^{n}}]$\end{tabular}}\\
(1,1,3) & \parbox[c]{2em}{
\includegraphics[scale=0.4]{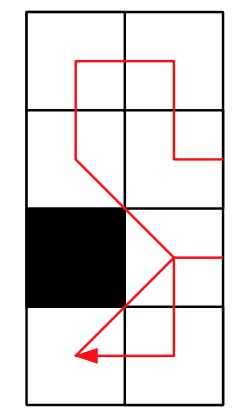}} &&\\ \hline

     \begin{tabular}{@{}c@{}}(0,1,0)\\(1,3,0) 
      \end{tabular}& \parbox[c]{2em}{
      \includegraphics[scale=0.35]{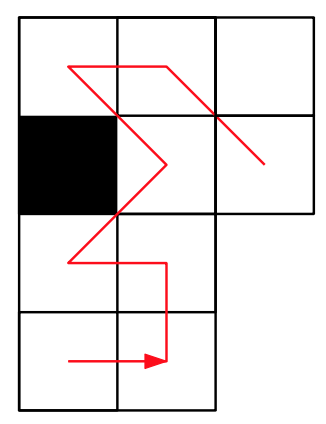}} & \multirow{2}{*}{\begin{tabular}{@{}c@{}}\\ \\ \\Node$(:,n_{obs})=$ Node$(:,n_{obs}+3)$\\Node$(:,n_{obs}+3)=$ Node$(:,n_{obs}+4)$\end{tabular}}& \multirow{2}{*}{\begin{tabular}{@{}c@{}} $(t-\frac{n_{obs}-1}{4^{n}})f_{h}(\frac{n_{obs}+3}{4^{n}})-(t-\frac{n_{obs}}{4^{n}})f_{h}(\frac{n_{obs}-1}{4^{n}})$\\ $\forall t \in [\frac{n_{obs}-1}{4^{n}},\frac{n_{obs}}{4^{n}}]$\\ \\$(t-\frac{n_{obs}}{4^{n}})f_{h}(\frac{n_{obs}+1}{4^{n}})-(t-\frac{n_{obs}+1}{4^{n}})f_{h}(\frac{n_{obs}-3}{4^{n}})$\\ $\forall t \in [\frac{n_{obs}}{4^{n}},\frac{n_{obs}+1}{4^{n}}]$\\ \\$(t-\frac{n_{obs}+2}{4^{n}})f_{h}(\frac{n_{obs}+4}{4^{n}})-(t-\frac{n_{obs}+4}{4^{n}})f_{h}(\frac{n_{obs}+2}{4^{n}})$\\ $\forall t \in [\frac{n_{obs}+2}{4^{n}},\frac{n_{obs}+4}{4^{n}}]$\end{tabular}} \\
      (1,2,0) &\parbox[c]{2em}{
      \includegraphics[scale=0.35]{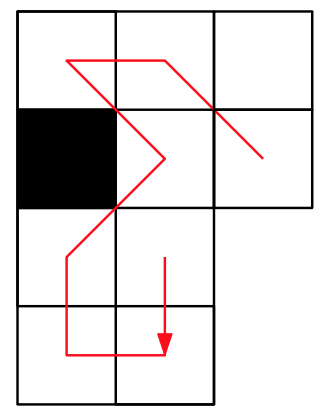}} &&\\

 \hline\hline
\end{tabular}
 \caption{Evasive maneuvers for obstacle avoidance.}
 \label{tabev}
\end{table*}


\begin{thm}
Using the evasion strategies enlisted in Table \ref{tabev}, a mobile agent traverses every available node of a Hilbert curve of any order in the space $\mathbf{Q}$ if an obstacle is placed on any node (barring the first and last node of the curve) of the same.\\
\end{thm}
\textit{Proof }
We proceed to prove the above statement using induction on the order of the Hilbert curve.\\
Consider the first order Hilbert curve. Since the obstacle can't be placed on the starting or ending node, it has to lie on a non-corner node as shown in figure \ref{nc}. Hence, the map is modified to just skip the blocked node and the rest of the nodes are visited. Hence, the proposed hypothesis holds true for $n=1$.\\\\
Suppose that the hypothesis holds true for $n=k$. Now, the Hilbert curve of order $n=k+1$ is composed of 4 units of $kth$ order Hilbert curves. There are two possible cases for the position of the obstacle node:\\
\textit{Case 1:} The obstacle node occupies a node which is neither the first nor the last node of one of the four $kth$ order Hilbert curves.\\
From definition \ref{def}, this would imply that obstacle doesn't lie on the corner nodes of $(k+1)th$ order Hilbert curve and the solution for $n=k$ is used for which the hypothesis holds true.\\\\
\textit{Case 2:} The obstacle node occupies a corner node of the $(k+1)th$ order Hilbert curve that isn't the starting or ending node of the same.\\
In this case, $(e_{n},p,m)$ are obtained and the corresponding maneuver is performed. Note that the evasive maneuvers preserve the sequence of nodes everywhere except the first order Hilbert curve within which the obstacle node lies. 
However, every node in this first order Hilbert curve is visited prior to resorting back to the correct sequence of nodes and thus every available node in the $(k+1)th$ order Hilbert curve is visited.\\

Since the hypothesis holds true for $n=k+1$ as well, we conclude using the principle of mathematical induction that using the evasion strategies enlisted above, a mobile agent traverses every available node of a Hilbert curve of any order in the space $\mathbf{Q}$ if an obstacle is placed on any node (barring the first and last node of the curve) of the same.\hspace{7.5cm}$\blacksquare$\\
\begin{cor}
The evasive strategies proposed in Table \ref{tabev} can accommodate multiple obstacles nodes provided that no two obstacles share an edge of their respective sub-squares.\\
\end{cor}
This modification of Hilbert's curve can be used online for exploration tasks by robotic agents. A specific example that could draw benefits from this algorithm is using ground bots for demining abandoned mine fields. After locating a search space, an array of ground bots can be left to explore the area and detonate the mines. Due to the modification of the map of the Hilbert's curve to accommodate for obstacles, detonated bots act as blocked nodes and are accordingly avoided by the remaining bots.
\section{Non-uniform Coverage}\label{S4}
Non-uniform coverage using a Hilbert's space filling curve has been studied in \cite{c8}. The region to be searched is divided into regions of varying ``interest". The ``interesting" regions are to be searched with a higher resolution (more finely, using a higher order Hilbert's curve). This is done by searching the space using coverage trees (\cite{c14}) where the root node is the centre of the square space and the subsequent levels are the nodes of Hilbert's curves of increasing order. Such a tree structure requires the use of a self-similar, locality preserving curve because when zooming in or out, the successive nodes need to be close to each other(\cite{c12}). This neccesitates the use of a space-filling fractal like the Hilbert's curve. Moreover, it is shown in \cite{c8} that using the Hilbert's curve for ordering of nodes in the coverage tree leads to more efficient coverage solutions. We extend their algorithm to accommodate for spaces with obstacles using the evasion strategy presented earlier (Table \ref{tabev}) via a simple modification. For the Shortcut Heuristic proposed in \cite{c14}, when the next node belongs to a Hilbert's curve of different order, the we first go to the parent or child node of the current node to match the order of the next node. Then, we proceed to the next node unless if it is an obstacle node; in which case, the proposed obstacle avoidance strategy can be used because the current node (after ascending or descending accordingly) and the obstacle node are of the same order. This modified strategy can also be used to avoid multiple obstacle nodes of orders different from that of the nodes dictated by required search resolution. The implicit assumption is that the obstacles are nodes of Hilbert's space-filling curve and it is also assumed that no two obstacles are adjacent.
\begin{figure}[h]
\centering
\includegraphics[scale=0.2]{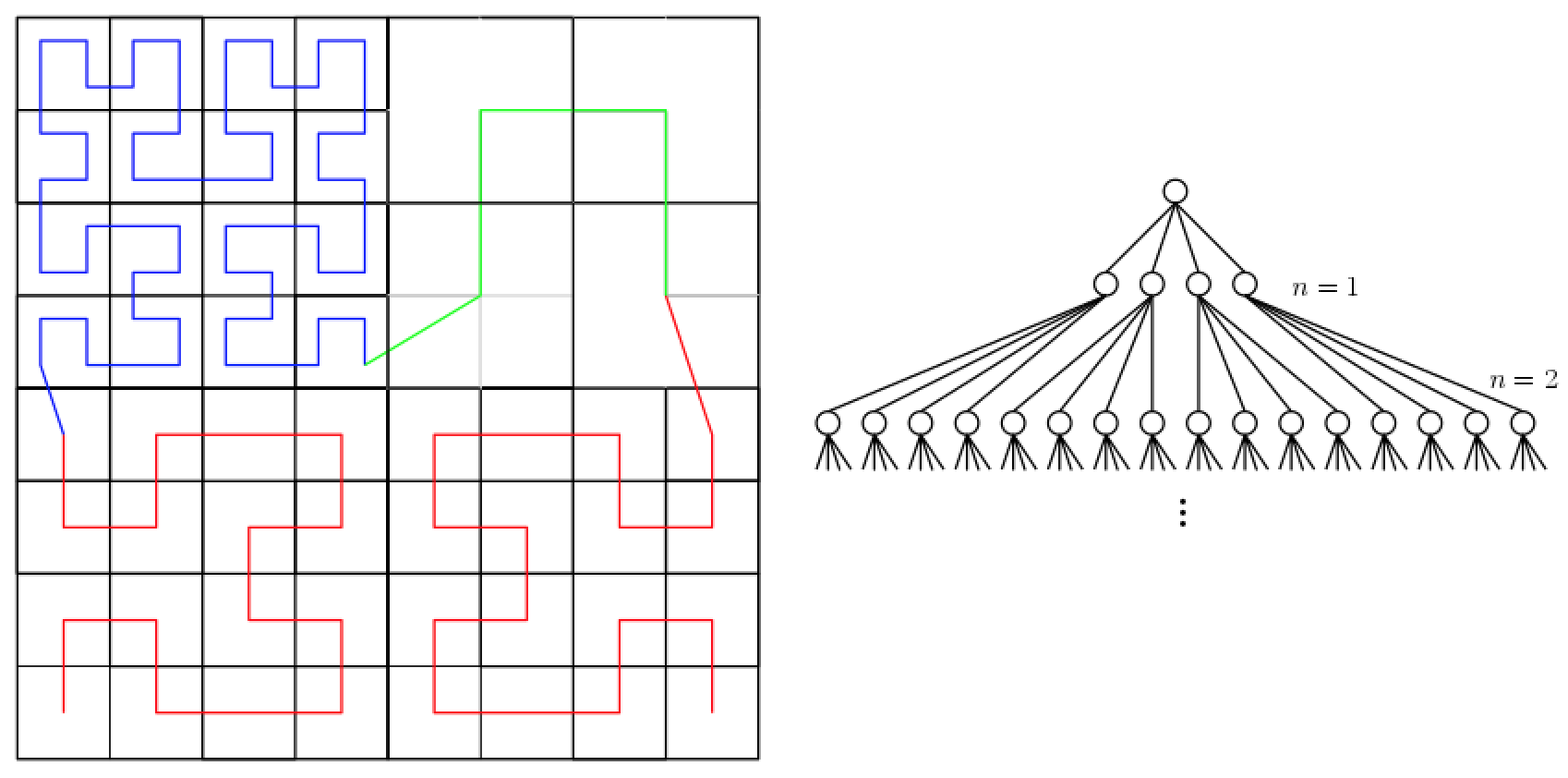}
\caption{Using Hilbert's curves for ordering of nodes in the Coverage Tree for non-uniform coverage}
\label{e}
\end{figure}

\section{Implementation and Simulations}\label{S5}
\begin{figure}[H]
\centering
  \begin{subfigure}[l]{0.25\textwidth}
  \centering 
   \includegraphics[scale = 0.13]{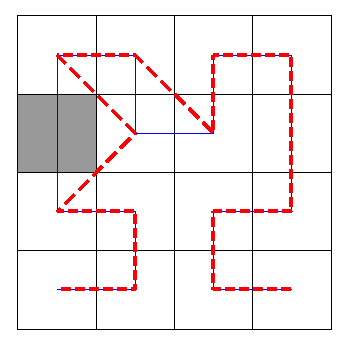}
    \caption{$(e_{n},p,m)=(0,0,0)$}
  \end{subfigure}\begin{subfigure}[l]{0.25\textwidth}
   \centering
    \includegraphics[scale = 0.13]{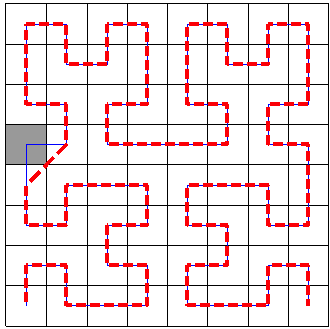}
    \caption{$(e_{n},p,m)=(1,0,0)$}
  \end{subfigure}

  \begin{subfigure}[l]{0.25\textwidth}
  \centering  
   \includegraphics[scale = 0.13]{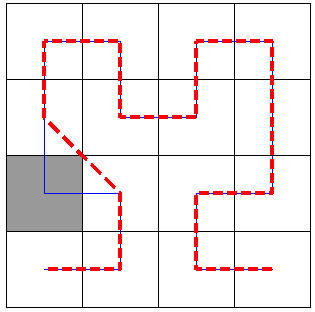}
    \caption{$(e_{n},p,m)=(0,0,3)$}
  \end{subfigure}\begin{subfigure}[l]{0.25\textwidth}
   \centering
    \includegraphics[scale = 0.13]{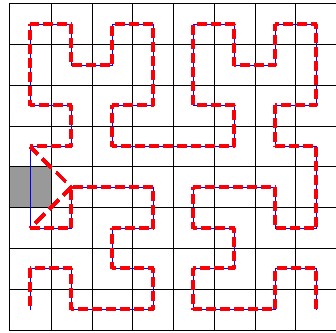}
    \caption{$(e_{n},p,m)=(1,0,3)$}
  \end{subfigure}
  
  \begin{subfigure}[l]{0.25\textwidth}
  \centering
   \includegraphics[scale = 0.12]{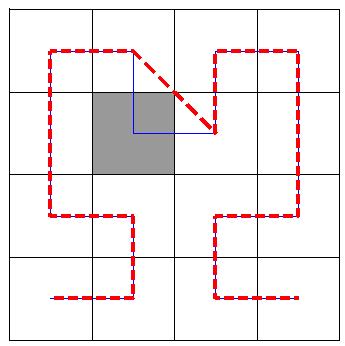}
    \caption{$(e_{n},p,m)=(0,1,3)$}
  \end{subfigure}\begin{subfigure}[l]{0.25\textwidth}
   \centering
    \includegraphics[scale = 0.12]{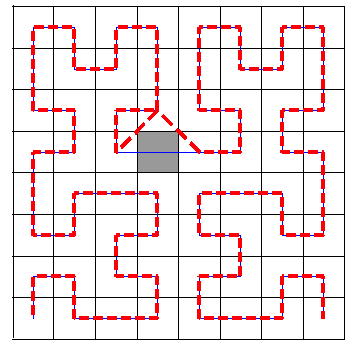}
    \caption{$(e_{n},p,m)=(1,1,3)$}
  \end{subfigure}
  \caption{}
  \label{sim}
\end{figure}

The algorithm presented in the earlier sections can be used for online implementation because no apriori information about the location of the obstacle nodes is required. For non-uniform coverage, the resolutions with which particular sections of the space are to be searched can be relayed to a ground agent by an aerial agent working in tandem in real time. The prescribed strategies are deployed in various situations to demonstrate its effectiveness in figure \ref{sim}. The various cases are tested for $n=2 (e_{n}=0)$ and $n=3 (e_{n}=1)$ for both types of corner nodes for the cases $p=0,1$.

\begin{algorithm}[h]
Repeat for every node\\
 \textbf{Input\ {$n$, $node\_is\_obstacle$, $obstacle\_size$}}\\
 \textbf{Memory\ {$t$}}\\
  $t=t+\frac{1}{4^{n}}$;\\
 Node$(:)=\begin{bmatrix}
 0\\0
 \end{bmatrix}$; \%\% Initializing vector of nodes \\
  $q(1:n)=[0 0..0]$; \%\%Vector to store digits\\
  $j=n$;\\
   \%\% Extracting digits\\
  \textbf{While\{$j\geq 1$\}\{\\}
  \quad$q(j)=4(4^{j-1}t-floor(4^{j-1}t))-4^{j}t+floor(4^{j}t)$\\
  \quad$j=j-1$;\}\\
  \%\% Mapping node via the Modified Hilbert Map\\
  $k=n$;\\
  \textbf{While\{$k\geq 1$\}\{\\}
\textbf{ $\quad$ if \{$k=n$\}\{\\}
  \quad\quad Node$(:)=F(q(n))$;\}\\
  \quad\textbf{else\{\\}
    \quad\quad Node$(:)=T_{q(k)}$Node$(:)$;\}\\
  \quad $k=k-1$;\}\\
  
 \%\%Obstacle Avoidance\\
 \textbf{if {$node\_is\_obstacle$}}\{\\
  \textbf{if} {Node$(:)$ is a corner node}\{\\
 $n=obstacle\_size$;\\
  Calculate Node(:) at $t-\frac{3}{4^n}$, $t$, $t+\frac{1}{4^n}$, $t+\frac{2}{4^n}$, $t+\frac{3}{4^n}$, $t+\frac{4}{4^n}$;\\
  Obtain $(e_{n_{eff}}, p, m)$ at $t$ and perform evasion maneuver;\}\\
  \textbf{else\{\\}
  Not a corner node, so just skip the node;\}\}\\
  \textbf{else\{\\}
  Make the robot go to the node;\}\\

%
 \caption{Implementation of prescribed algorithm.}
\end{algorithm}


%
%
%

\begin{figure}[H]
\centering
\begin{subfigure}{.25\textwidth}
\captionsetup{width=0.8\textwidth}
  \centering
  \includegraphics[scale=0.3]{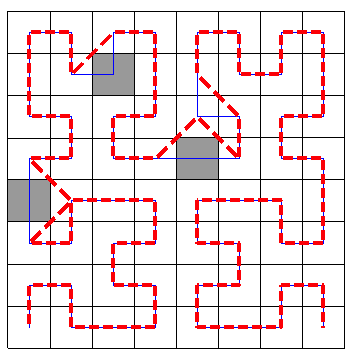}
  \caption{Employing enlisted strategies in a case with multiple obstacles.}
 \label{m}
\end{subfigure}\begin{subfigure}{.25\textwidth}
\captionsetup{width=0.8\textwidth}
  \centering
  \includegraphics[scale=0.08]{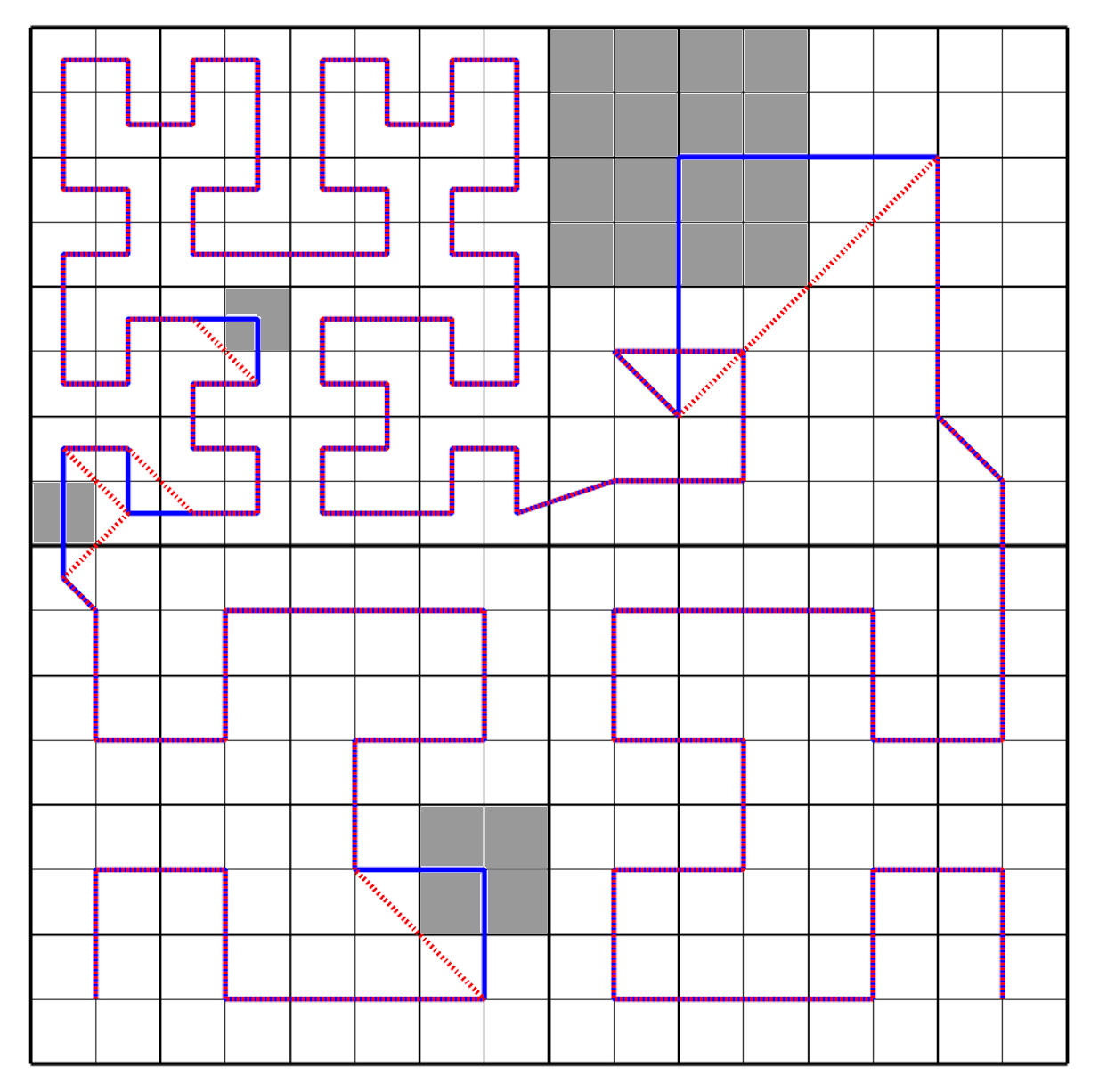}
  \caption{Non-uniform coverage in a space with obstacles.}
\label{e}
\end{subfigure}
\caption{}
\end{figure}

If the obstacles are far enough, the prescribed evasive strategies yield good results as shown in Figure \ref{m}. Cases like this could appear in real life situations like demining of mine fields as mentioned earlier. Furthermore, application of the algorithm for non-uniform coverage of a space with obstacles of varying sizes is presented in Figure \ref{e}.

\section{Conclusion}\label{S6}
The paper suggested a change in the map for plotting Hilbert's curve via its approximating polygon for use in exploration tasks. In the following sections, the problem of using the Hilbert's curve for exploring a space with an obstacle/a hole was considered and an inductive solution to the same was obtained using the locality preserving and self-similarity property of the Hilbert curve. The prescribed strategy is suitable for online implementation because location of the obstacle is not required apriori. The algorithm can be also used during non-uniform coverage operations. Future work entails consideration of arbitrary obstacles occupying a space to develop strategies for path planning and searching.


\bibliographystyle{IEEEtran}
\bibliography{IEEEabrv,IEEEexample}

\end{document}